\pdfoutput=1

\documentclass[11pt]{article}

\usepackage[final]{acl}

\usepackage{times}
\usepackage{latexsym}
\usepackage{bbm} 
\usepackage[T1]{fontenc}

\usepackage[utf8]{inputenc}

\usepackage{microtype}

\usepackage{inconsolata}
\usepackage{microtype}
\usepackage{graphicx}
\usepackage{import}
\usepackage{adjustbox}
\usepackage{layout}
\usepackage{tabularx, makecell}
\usepackage{booktabs}
\usepackage{mathrsfs}
\usepackage{amssymb} 
\usepackage{url}
\usepackage{hyperref}
\usepackage{graphicx}
\usepackage{xspace,paralist}
\usepackage{times,latexsym}
\usepackage{amsmath}
\usepackage{appendix}
\usepackage{comment} 
\usepackage{enumitem}
\usepackage{makecell}
\usepackage{multirow}
\usepackage{xcolor}
\usepackage{arydshln}
\usepackage{cleveref}
\usepackage{todonotes}
\usepackage{longtable,supertabular}
\usepackage{amssymb}
\usepackage{pifont}
\usepackage{CJKutf8}
\usepackage{subcaption}
\usepackage{ragged2e}

\NewDocumentCommand{\revanth}
{ mO{} }{\textcolor{blue}{\textsuperscript{\textit{Revanth}}\textsf{\textbf{\small[#1]}}}}
\title{One More Turn, Less Regret:\\ A Regret-Based Multi-Turn Benchmark for LLMs' Clarification Policies}

\author{Minh Ngoc Ta\textsuperscript{1},
My Anh Tran Nguyen\textsuperscript{2},
Duong D. Nguyen\textsuperscript{2},
Yuxia Wang\textsuperscript{3},
Preslav Nakov\textsuperscript{1}
\\
\textsuperscript{1}MBZUAI \quad
\textsuperscript{2}BKAI Research Center, Hanoi University of Science and Technology \\
\textsuperscript{3}INSAIT, Sofia University "St. Kliment Ohridski"
\\ \\
{\texttt{\{minh.ta, preslav.nakov\}@mbzuai.ac.ae}}
\\
}

\begin{document}
\maketitle

\begin{abstract}
Ambiguous user requests make clarification a sequential decision problem for conversational LLM assistants: they must decide whether to ask, what to ask, when to stop, and when to answer. We introduce \emph{RegretBench}, a multi-turn benchmark that evaluates clarification as policy behavior rather than isolated question quality. RegretBench provides a hidden-intent formulation of ambiguity, supports free-form interaction grounded in semantic-state tracking, and introduces a regret-based objective that measures how much value a model loses relative to a reference clarification policy. Experiments on open-domain QA and product recommendation scenarios show that final success alone is insufficient, as models with similar accuracy can differ substantially in efficiency, robustness to user behaviors, and stopping decisions. By jointly measuring intent resolution, interaction cost, ineffective clarification, and regret, RegretBench reveals whether models clarify usefully and efficiently. Our results show that effective clarification requires more than plausible questions: models must ask the right question at the right time and stop once the user’s intended meaning is clear.
\end{abstract}

\section{Introduction}

Large language models (LLMs) are increasingly used as conversational assistants, but users often ask questions without fully specifying what they mean~\cite{keluskar2024llms}. In these settings, the challenge is not only how to phrase a response, but whether to respond at all. It must decide whether to answer, ask a clarification question, continue asking follow-ups, or stop clarifying and commit to an answer. If the assistant answers too early, it may misunderstand the problem~\cite{abe2025llm}. If it asks for too long, the interaction becomes costly for the user. A good assistant must infer what the user likely means and decide whether another question would help enough to justify asking it.

The main uncertainty is the user's intended meaning. The same query can point to more than one reasonable interpretation~\cite{keluskar2024llms}. The assistant model should ask when the user's meaning is still unclear, and answer once it has enough information. Clarification is therefore more than question generation. The assistant must decide what missing detail to ask about and whether the answer is worth the extra interaction.

A growing line of work has studied clarification in information-seeking dialogue. Datasets such as Qulac~\cite{aluric2020qulac} and ClariQ~\cite{aliannejadi2020clariq} focus on clarifying question selection and ranking, while AmbigQA~\cite{min2020ambigqa} and CAmbigNQ~\cite{lee2023cambignq} expose multiple valid interpretations of ambiguous questions and evaluate downstream question answering. More recent benchmarks, including CLAMBER~\cite{zhang2024clamber}, InfoQuest~\cite{infoquest}, and ClarifyMT-Bench~\cite{clarifymt}, broaden ambiguity evaluation to richer ambiguity types and multi-turn interactions. \citet{zhang-choi-2025-clarify} further emphasizes that assistants must decide when clarification is warranted, rather than always asking or always answering.

Despite this progress, existing evaluations still leave a fundamental gap. Many benchmarks evaluate local capabilities: whether a single clarification question is relevant, whether ambiguity is detected, whether an answer is correct after a fixed interaction, or whether a transcript is reasonable after the fact \cite{chen-etal-2024-style,shi-etal-2025-ambiguity, gan2024clarqllmbenchmarkmodelsclarifying}. These signals are useful, but they still miss the main question: whether the assistant identifies the user's goal without unnecessary interaction. A model can ask a reasonable-sounding question that does not help. It may keep asking after the intent is clear, answer too early, or work well only when users give clean replies. Such failures are hard to capture with per-turn metrics or fixed-transcript evaluation.

We introduce \textbf{RegretBench}, a regret-based benchmark for evaluating LLM clarification policies under hidden user intent. RegretBench treats the assistant as a dialogue policy acting in a partially observed environment. Each instance specifies an ambiguous prompt, a set of latent intents, and semantic variables that distinguish those intents. The evaluated model interacts in free-form natural language, while the benchmark runtime grounds the dialogue into a semantic interface for tracking intent's beliefs and evaluating decisions.

This design enables policy-level evaluation without requiring models to choose from a fixed question bank. Free-form clarification questions are mapped to semantic actions, user replies are simulated from hidden intent and persona, and the full dialogue is scored. Hence, RegretBench measures not only whether the model identifies the correct intent, but also the efficiency and robustness.

The core evaluation signal is whole-dialogue return, combining final intent correctness, interaction cost, and penalties for ineffective clarification. We report interpretable metrics such as intent success rate and average turns, plus regret relative to a semantic reference planner over the same benchmark state and action space. This regret measures benchmark-relative policy suboptimality: how much value a model loses compared with a strong reference strategy, without claiming an oracle optimum over arbitrary raw-language dialogue.

RegretBench is constructed by reusing existing ambiguity-focused resources as intent scaffolds and converting them into semantic clarification instances. This provides the benchmark with an explicit latent structure while preserving free-form interaction with the evaluated model. 

Our contributions are as follows:
\begin{itemize}
    \item We reformulate clarification evaluation as policy evaluation under hidden intent, making intent resolution the central objective rather than isolated clarification-question quality.

    \item We introduce RegretBench, a free-form multi-turn benchmark in which LLMs interact with a persona-conditioned user simulator, grounding dialogue in semantic actions and observations for intent tracking.

    \item We define a whole-dialogue reward that jointly captures final intent correctness, interaction efficiency, and ineffective or unsupported clarification behavior.

    \item We propose regret against a semantic reference planner as a benchmark-relative measure of clarification policy quality, enabling comparison beyond final-answer accuracy alone.
\end{itemize}

\section{Background and Related Work}

\subsection{Clarification and Ambiguity in Dialogue}

Clarification has been studied in information-seeking dialogue and conversational search, where users often issue short, underspecified, or ambiguous requests. Early benchmarks focused on generating, selecting, or ranking useful clarifying questions. Qulac~\cite{aluric2020qulac} introduced an offline framework for asking clarifying questions in open-domain information-seeking conversations, and ClariQ~\cite{aliannejadi2020clariq} extended this research topic through shared tasks on clarification question generation and ranking. These works established clarification as an important capability for interactive systems, but their evaluation is primarily local: a model is assessed on whether it produces or selects a good clarification question at a particular step.

Another relevant direction examines ambiguity in open-domain question answering. AmbigQA~\cite{min2020ambigqa} makes ambiguity explicit by explaining a single question with multiple plausible interpretations, rewrites, and answers. CAmbigNQ~\cite{lee2023cambignq} brings ambiguous clarification QA closer to dialogue by adding clarification questions before the final QA target. These datasets are useful because they make the different possible readings of a question explicit. However, they are typically evaluated through component tasks such as ambiguity detection, clarification generation, rewrite prediction, or final QA accuracy. They do not directly evaluate the full interaction policy an assistant must execute: deciding whether to ask, what to ask, how to use the user's reply, and when to stop.

More recent benchmarks broaden the evaluation of ambiguity for LLMs.  CLAMBER~\cite{zhang2024clamber} introduces a taxonomy-driven benchmark for recognizing and clarifying ambiguous information needs across diverse ambiguity types. Such work highlights that ambiguity can arise from many sources, including entities, attributes, temporal references, scope, and task constraints. Nevertheless, the main evaluation object remains largely instance-level: whether ambiguity is detected and whether a plausible clarification is produced.

RegretBench builds on this line of work but shifts the focus from individual clarification quality to the entire clarification policy.

\subsection{Toward Multi-Turn Clarification Policies}

A closely related direction is to study when clarification is necessary. \citet{zhang-choi-2025-clarify} framed clarification as a cost-sensitive decision, emphasizing that asking a question is useful only when the expected benefit outweighs the burden of additional interaction. This perspective is central to realistic assistant behavior: a good model should neither always ask nor answer prematurely when the user's intent remains uncertain.

However, the ask-versus-answer decision is only one part of the broader clarification problem. Choosing to clarify is only the first step. After deciding to clarify, the model still has to do more than ask a follow-up. Even after choosing to clarify, the hard part is not over. The model has to find the question that actually help, make something useful of the reply, and know when to stop asking. Recent benchmarks such as InfoQuest~\cite{infoquest} and ClarifyMT-Bench~\cite{clarifymt} also study multi-turn clarification with hidden information, diverse ambiguity types, and simulated user responses. These benchmarks are important steps toward realistic evaluation, but they are not explicitly designed as policy-level evaluations over a defined latent intent state, semantic action space, whole-dialogue reward, and reference planner.

RegretBench is designed around this missing policy-level view. It treats the assistant as a dialogue policy under hidden user intent. The evaluated model interacts in free-form natural language, while the benchmark runtime maps clarification questions to semantic ask actions and user replies to structured observations. This makes it possible to track whether a dialogue actually reduces uncertainty about the user's intended meaning, rather than merely producing reasonable-sounding turns.

\subsection{Grounded Evaluation of Clarification}

Open-ended dialogue evaluation often relies on model-based judgment. Benchmarks such as MT-Bench~\cite{zheng2023judging} show that LLM judges can be useful for comparing conversational responses when exact-match metrics are too restrictive. Meanwhile, later work has shown that judge-based evaluation can be brittle and sensitive to the judge model, prompt, and evaluation setup~\cite{lee-etal-2025-ember}.

These issues are especially important for clarification: a question may sound natural and helpful while failing to distinguish the relevant intents, or a dialogue may appear fluent while resolving the wrong ambiguity.

RegretBench, therefore, does not use an LLM judge as its core scoring mechanism. Instead, it grounds evaluation in benchmark-hidden intent. Each instance defines an ambiguous prompt, a set of latent intents, semantic variables that distinguish those intents, supported clarification actions, and structured observations generated by a persona-conditioned simulator. The final score is computed over the whole dialogue, combining intent success, interaction cost, penalties for unsupported or ineffective clarification behavior, and regret relative to a semantic reference planner.

This positioning connects prior work on clarification question generation, ambiguity-aware QA, when-to-clarify decisions, and multi-turn interaction, while addressing a different evaluation target. RegretBench evaluates whether an assistant can efficiently and robustly recover the user's true intent across an entire dialogue, including when to ask, what to ask, and when to stop. The goal is not only to ask plausible clarification questions, but to make good sequential decisions under uncertainty that lead to successful intent resolution.
\section{Data Formulation}
\label{sec:dataset}
RegretBench represents each ambiguous user request as a clarification interaction graph (CIG). A CIG specifies the candidate intents, the latent variables that distinguish them, and the semantic actions and observations used to resolve uncertainty. This lets models interact in free-form natural language while the benchmark evaluates them in an explicit semantic state space.

\subsection{Clarification Interaction Graph}

For a prompt $x$, a CIG defines a set of latent intents $Z$, latent variables $V$, a semantic action space $\mathcal{A}$, and a semantic observation space $\mathcal{O}$:
\begin{equation}
G_x = \left(x, Z, V, \mathcal{A}, \mathcal{O}\right).
\end{equation}

Table~\ref{tab:cig_fields} explains the core components of a CIG. The prompt $x$ is the user request shown to the assistant. Each intent $z \in Z$ is one plausible interpretation, represented by an identifier, a natural-language description, and slots. Slots include answer aliases for final response evaluation and values for the latent variables that distinguish intents.

Latent variables capture the ambiguity dimensions. For example, if a prompt refers to either a male or female contestant, the CIG may include \texttt{gender} with intent-specific values. Each retained variable induces a supported semantic ask action, such as \texttt{ask:gender}. The model does not see these actions directly; it asks free-form questions, and the runtime maps each question to the closest supported action or marks it as unsupported when it does not target a valid ambiguity dimension.


The observation schema defines how the simulated user can respond to a clarification question. A valid ask can produce an informative reply, such as \texttt{provide\_slot:gender:male}, updating belief toward intents with that slot value. Depending on the persona, the simulator may also return less informative outcomes such as \texttt{uncertain}, \texttt{contradict}, or \texttt{off\_topic}. Unsupported questions are marked as \texttt{unsupported}, and do not update belief. Thus, the CIG specifies how dialogue turns generate evidence about the hidden intent, enabling belief tracking, user simulation, reward computation, and reference planning.

\begin{table}
\centering
\small
\begin{tabular}{p{0.11\linewidth}p{0.76\linewidth}}
\toprule
\textbf{Field} & \textbf{Description} \\
\midrule
$x$ & Ambiguous natural-language request shown to the evaluated assistant. \\
$Z$ & Hidden candidate interpretations of the prompt, each with a description and slot values. \\
$V$ & Semantic dimensions that distinguish the candidate intents. \\
$\mathcal{A}$ & Supported semantic clarification actions used by the runtime and reference planner. Each action corresponds to one ambiguity variable and is associated with one or more reference questions for grounding free-form questions. \\
$\mathcal{O}$ & Structured user observations used for simulation and belief updates. \\
\bottomrule
\end{tabular}
\caption{Core field description for a CIG $G_x$.}
\label{tab:cig_fields}
\end{table}

This design deliberately separates the surface form of interaction from the semantic structure used for evaluation. The model is not constrained to select from a fixed question bank. However, we can still evaluate the model's questions based on whether they target a valid ambiguity variable and reduce uncertainty about the hidden intent.

\subsection{Dataset Collection and Preparation}

\begin{table*}
\centering
\resizebox{\textwidth}{!}{
\begin{tabular}{p{0.1\linewidth}p{0.1\linewidth}llrrlp{0.3\linewidth}}
\toprule
\textbf{Subset} & \textbf{Dataset} & \textbf{Creator} & \textbf{License} & \textbf{Size} & \textbf{Used} & \textbf{Domain} & \textbf{Main ambiguity structure} \\
\midrule
Open-domain QA & AmbigDocs & \citet{ambigdocs} & Apache 2.0 & 7,220 & 4,836 & Document QA 
& Multiple document-backed referents for the same ambiguous surface form. \\
& CondAmbi-gQA-2K & \citet{li-etal-2025-condambigqa} & Apache 2.0 & 2,000 & 1,451 & Conditional QA 
& Different answers under different explicit conditions. \\
\midrule
Real-world use case
& PSCon & \citet{zou2025pscon} & CC-BY-4.0 & 1,730 & 377 & \makecell[lt]{Product search /\\ recommendation} & Multiple candidate products compatible with the same user request. \\
\bottomrule
\end{tabular}
}
\caption{RegretBench CIGs subsets creation after source-level filtering and normalization.}
\label{tab:dataset_sources}
\end{table*}


We construct RegretBench from two subsets: \emph{(i)}~open-domain QA, covering document-grounded entity ambiguity and condition-dependent question answering, and \emph{(ii)}~real-world use case, covering conversational product recommendation. 

The source datasets are summarized in Table~\ref{tab:dataset_sources}. Because these datasets use different formats, we first normalize their records into a shared ambiguous-QA schema consisting of a user-facing primary question and a set of candidate intents. Each intent includes a context or description together with one or more answer aliases. We then convert each record into a CIG: the primary question becomes the prompt, each interpretation becomes a hidden intent, and the generator infers the remaining structure from the supporting context.

A key design choice is that source documents are used only for curation and benchmark construction. At test time, the assistant sees only the user question and dialogue history, not the ambiguous entity annotation, document titles, snippets, candidate answers, or hidden intents. We keep only examples where the need for clarification is visible from the user question itself. This focuses RegretBench on ambiguity handling rather than obscure trivia, hidden-document lookup, or external recall.

\paragraph{AmbigDocs~\cite{ambigdocs}.}
AmbigDocs pairs ambiguous questions with candidate documents, each representing a possible referent with a title, snippet, and target answer. We retain examples with clean, realistic alternatives: titles must denote genuine referents of the same phrase, and snippets must directly support concrete, distinct answers. We map the original question to the primary question, each candidate document to one intent, and each snippet to the intent context.

\paragraph{CondAmbigQA-2K~\cite{li-etal-2025-condambigqa}.}
It contains 2,000 ambiguous queries with condition-answer-citation annotations. We keep records with at least two valid conditional interpretations. The original question becomes the primary question; each condition-specific answer becomes one intent; and the condition plus ground truth forms the intent context. Compact answer aliases are derived by extracting concise core spans from the ground truth.

\paragraph{PSCon~\cite{zou2025pscon}.}
To cover ambiguity beyond QA, we adapt PSCon for product search. We use 377 English records with more than two candidate products. The user request becomes the primary question, candidate products from search results serve as intent contexts, and answers are derived by matching product keywords to product-rate annotations. These examples test whether assistants can clarify user preferences before making a recommendation.

\paragraph{Filtering and validation.}
Before CIG generation, we apply a precision-oriented filter. We keep records only when the ambiguity is recognizable from the question, candidate referents are cleanly distinguishable, no single referent overwhelmingly dominates, and clarification would change the final answer. We remove obscure biography traps, tiny-place lookup artifacts, hidden-document-only ambiguity, unnatural wording, poor candidate sets, unsupported answers, and already-resolved referents. Borderline cases are dropped to prioritize quality over coverage. Detailed criteria and examples are provided in Appendix~\ref{app:dataset_filtering}.
\section{Metrics}

RegretBench evaluates clarification as a whole-dialogue policy problem under hidden user intent. Instead of scoring isolated question quality or per-turn accuracy, it measures whether the assistant identifies the user's true intent, how much interaction cost it incurs, and whether its clarification behavior is grounded. All metrics are computed under a fixed evaluation configuration, including split, persona, horizon, rollout count, simulator backends, random seed, and cost profile. The main metrics are reward, regret, normalized regret, and goal success rate; additional diagnostics explain model behavior.

\subsection{Reward Function}

RegretBench uses a whole-dialogue reward to capture final intent correctness, interaction cost, and ineffective clarification behavior. For each rollout, let $z$ be the hidden user intent, $\tau$ the dialogue trajectory, and $\hat{z}$ the final intent selected by the assistant. The rollout reward is:
\begin{equation}
R(\tau, z) = U(\hat{z}, z) - C(\tau) - P(\tau)
\end{equation}
where $U(\hat{z}, z)$ is terminal goal utility, $C(\tau)$ is interaction cost, and $P(\tau)$ is the accumulated penalty.

The terminal utility is an exact intent match:
\begin{equation}
U(\hat{z}, z) =
\begin{cases}
1, & \text{if } \hat{z} = z \\
0, & \text{otherwise}
\end{cases}
\end{equation}

The cost profile uses assistant-side token cost:
\begin{equation}
C(\tau) = \lambda \times \sum_t N^{\mathrm{asst}}_t
\end{equation}
where $N^{\mathrm{asst}}_t$ is the number of assistant-side tokens at turn $t$, and $\lambda$ is a fixed normalized token-cost weight. Benchmark-side tokens from the simulator are not charged in the primary reward.

The penalty term is:
\begin{equation}
P(\tau) = \sum_t p_t
\end{equation}
Penalized events include unsupported clarification actions, uncertain or off-topic user outcomes, contradictions, refusals, redundant clarification after the intent is effectively resolved, and forced terminal answers when the model fails to commit before the horizon.

For each instance $x$, the hidden intent is sampled uniformly from its candidate intent set $Z_x$. The instance-level reward of policy $\pi$ is:
\begin{equation}
\mathcal{R}_x(\pi)
=
\frac{1}{\left|Z_x\right|}
\sum_{z \in Z_x}
\mathbb{E}_{\tau \sim \pi,\,\mathrm{sim}(\cdot \mid x,z)}
\left[
R(\tau, z)
\right]
\end{equation}

The dataset-level reward is:
\begin{equation}
\mathcal{R}(\pi)
=
\frac{1}{\left|D\right|}
\sum_{x \in D}
\mathcal{R}_x(\pi)
\end{equation}

This is the main absolute performance score: a strong policy should identify the correct intent, ask only useful clarification questions, avoid unsupported actions, and stop once enough information has been obtained.

\subsection{Regret}

Dataset-level rewards are absolute, but instances differ in ambiguity and the value of clarification. RegretBench therefore compares each policy to a semantic reference planner operating over the same intent space, semantic actions, observation model, cost function, and horizon.

For instance,
\begin{equation}
\mathrm{Regret}_x(\pi)
=
\mathcal{R}_x(\pi_{\mathrm{ref}}) - \mathcal{R}_x(\pi)
\end{equation}
where $\pi_{\mathrm{ref}}$ is the benchmark reference planner. Dataset-level regret is
\begin{equation}
\mathrm{Regret}(\pi)
=
\dfrac{1}{\left|D\right|}
\sum_{x \in D}
\mathrm{Regret}_x(\pi)
\end{equation}

Regret measures the amount of whole-dialogue value a policy loses relative to the reference strategy. High regret can indicate premature answering, uninformative or redundant questions, unsupported clarification, or failure to commit. Since the planner is defined over the benchmark semantic space, regret should be interpreted as benchmark-relative policy suboptimality, not as optimality over arbitrary raw-language dialogue.

\subsection{Normalized Regret}

Normalized regret scales raw regret by the amount of value clarification it can provide on each instance:
\begin{equation}
\mathrm{N.Regret}_x(\pi)
=
\frac{
\mathrm{Regret}_x(\pi)
}{
\max\left(
\mathcal{R}_x(\pi_{\mathrm{ref}}) - \mathcal{R}_x(\pi_{\mathrm{base}}),
\varepsilon
\right)
}
\end{equation}

Here, $\pi_{\mathrm{base}}$ is the best immediate-answer baseline, and $\varepsilon$ is a small constant for numerical stability. The denominator is the available clarification margin, namely, how much the reference planner improves over answering immediately. Values near zero indicate behavior close to the reference planner; values near one indicate performance close to the immediate-answer baseline; values above one indicate worse-than-baseline performance under the benchmark reward model.

\subsection{Goal Success Rate}

Goal success rate is the fraction of rollouts in which the assistant's final selected intent matches the hidden user intent:
\begin{equation}
\mathrm{SuccessRate}
=
\dfrac{1}{N}
\sum_{i=1}^{N}
\mathbbm{1}\!\left[\hat{z}_i = z_i\right].
\end{equation}

This is the most direct measure of final correctness. It should be read together with reward and cost: high success with many turns may indicate over-clarification, while low success with few turns may indicate premature answering.

\subsection{Supporting Metrics and Diagnostics}

RegretBench also reports supporting diagnostics, including average turns, cost, penalty, unsupported clarification rate, and forced terminal answer rate. These metrics are used for interpretation rather than primary ranking: they indicate whether a model succeeds through efficient, grounded clarification or fails through over-asking, unsupported questions, or poor stopping behavior. Full definitions are provided in Appendix~\ref{app:supporting_metrics}.

\section{Experiments} 

\begin{table*}
\centering
\resizebox{\textwidth}{!}{
\begin{tabular}{ll|rrrr|rrr}
\hline
\textbf{Model} & \textbf{Publisher}& \textbf{Reward} $\uparrow$ & \textbf{Regret} $\downarrow$ & \textbf{N.Regret} $\downarrow$ & \textbf{Success} $\uparrow$ & \textbf{Avg. Turn} $\downarrow$ & \textbf{Avg. Cost} $\downarrow$ & \textbf{Avg. Penalty} $\downarrow$\\
\hline
\multicolumn{8}{l}{\textit{Reasoning models}} \\
\hline
Gemini 3.1 Pro & \citet{gemini} & \textbf{0.6586} & \underline{0.3395} & \textbf{0.5625} & \textbf{0.8392} & 1.3806 & 0.1398 & 0.0409 \\
K2Think V2 & \citet{k2think} & $-$4.2355 & 5.2312 & 8.7523 & 0.7449 & 4.0759 & 4.5842 & 0.4762 \\
Grok 4.1 Fast & \citet{grok} & 0.5947 & 0.4034 & 0.6960 & 0.7958 & 1.1502 & 0.1310 & 0.0301 \\
GPT 5.4 Nano & \citet{gpt5} & 0.4840 & 0.5112 & 0.8858 & 0.7174 & 1.4518 & 0.1574 & 0.7607 \\
GPT 5 Mini & \citet{gpt5} & 0.5498 & 0.4456 & 0.7667 & 0.7744 & 1.1805 & 0.1838 & 0.0408 \\
Nemotron 3 Super 120B & \citet{nemotron} & 0.4605 & 0.5354 & 0.9276 & 0.7068 & 1.1459 & 0.1937 & 0.5263 \\
Qwen 3.5 Flash & \citet{qwen} & 0.5655 & 0.4301 & 0.7456 & 0.7110 & 1.0663 & 0.1136 & 0.0319 \\
DeepSeek V4 Flash & \citet{deepseek} & 0.6091 & 0.3861 & 0.6716 & 0.7320 & 1.0645 & \textbf{0.0994} & \underline{0.0235} \\

\hline
\multicolumn{8}{l}{\textit{Non-reasoning models}} \\
\hline
Gemini 3.1 Flash-Lite & \citet{gemini} & 0.4855 & 0.5107 & 0.8634 & 0.7501 & 1.6331 & 0.1688 & 0.0958 \\
K2-V2 & \citet{k2v2} & 0.2736 & 0.6230 & 1.0164 & 0.3913 & \textbf{0.0019} & 0.1777 & 0.1465 \\
Grok 4.1 Fast & \citet{grok} & 0.5671 & 0.4288 & 0.7294 & \underline{0.7991} & 1.3909 & 0.1784 & 0.0536 \\
GPT 4.1 & \citet{gpt41} & \underline{0.6584} & \textbf{0.3373} & \underline{0.5823} & 0.7930 & 1.1954 & 0.1134 & \textbf{0.0211} \\
Llama 4 Scout & \citet{llama4} & 0.4386 & 0.5568 & 0.9722 & 0.7952 & 2.1413 & 0.2193 & 0.1373 \\
Llama 4 Maverick & \citet{llama4} & 0.5649 & 0.4307 & 0.7449 & 0.7746 & 1.6321 & 0.1385 & 0.0712 \\
Qwen 3.5 Flash\footnotemark & \citet{qwen} & 0.4655 & 0.5298 & 0.9235 & 0.6391 & \underline{0.9580} & 0.1260 & 0.0475 \\
DeepSeek V4 Flash\textsuperscript{\ref{thinking}} & \citet{deepseek} & 0.6243 & 0.3715 & 0.6436 & 0.7536 & 1.1561 & \underline{0.1029} & 0.0264 \\
\hline
\end{tabular}
}
\caption{Performance across models. The best results are in \textbf{bold} and the second best are \underline{underlined}.}
\label{tab:exp-models}
\end{table*}

\begin{table*}
\centering
\resizebox{\textwidth}{!}{
\begin{tabular}{ll|rrrr|rrr}
\hline
\textbf{Model} & \textbf{Method} & \textbf{Reward} $\uparrow$ & \textbf{Regret} $\downarrow$ & \textbf{N.Regret} $\downarrow$ & \textbf{Success} $\uparrow$ & \textbf{Avg. Turn} $\downarrow$ & \textbf{Avg. Cost} $\downarrow$ & \textbf{Avg. Penalty} $\downarrow$\\
\hline
Qwen 3.5 Flash & -- & 0.5655 & 0.4301 & 0.7456 & 0.7110 & 1.0664 & 0.1136 & 0.0319 \\
Qwen 3.5 Flash & One-shot & 0.5707 & 0.4250 & 0.7356 & 0.6929 & \textbf{0.9452} & \textbf{0.1023} & \textbf{0.0199} \\
Qwen 3.5 Flash & Explanation & \textbf{0.5856} & \textbf{0.4101} & \textbf{0.7102} & \textbf{0.7147} & 1.0112 & 0.1068 & 0.0223 \\
\hline
Llama 4 Maverick & -- & 0.5649 & 0.4307 & 0.7449 & 0.7746 & 1.6321 & 0.1385 & 0.0712 \\
Llama 4 Maverick & One-shot & 0.5679 & 0.4277 & 0.7398 & 0.7699 & \textbf{1.0663} & \textbf{0.1136} & \textbf{0.0319} \\
Llama 4 Maverick & Explanation & \textbf{0.5754} & \textbf{0.4202} & \textbf{0.7267} & \textbf{0.7762} & 1.5982 & 0.1351 & 0.0658 \\
\hline
\end{tabular}
}
\caption{Performance with different clarification-oriented methods. The best results are in \textbf{bold}.}
\label{tab:exp-methods}
\end{table*}

\subsection{Experimental Setup}

All models are evaluated under the same setting: the cooperative persona, a maximum of five questioning turns, three rollouts, and an identical cost profile. This ensures that score differences primarily reflect the evaluated assistant policy rather than changes in the benchmark configuration. To maintain consistent user behavior, we use GPT-5 Nano as the user-reply model across all experiments.

\subsection{Which Models Behave Like Good Clarification Policies}

We evaluate all models under the same settings using the open-domain QA subset. Table~\ref{tab:exp-models} reports the main model comparison, while Table~\ref{tab:exp-methods} evaluates lightweight prompt-only clarification methods. Details of these methods are provided in Appendix~\ref{app:clarification_methods}.

Overall, the results show that high accuracy does not always imply a good clarification policy. As we can see in K2-Think V2, while its success rate is relatively high, its reward is strongly negative due to excessive turns, high costs, and substantial penalties. Conversely, GPT 4.1, a non-reasoning model, achieved a nearly optimal whole-dialogue reward and the lowest normalized regret of any model, only surpassed by Gemini 3.1 Pro. This shows that an explicit reasoning mode is not required to achieve strong performance in clarification.

This illustrates our central motivation: a model should not be evaluated just on whether it eventually reaches the correct intent, but on whether it asks useful questions efficiently and stops at the right time.

We see this pattern further reflected in Llama 4 Scout, which maintains a high success rate but a lower reward due to longer, less efficient interactions. In contrast, DeepSeek V4 Flash successfully navigates this balance, achieving strong rewards with low interaction costs, offering a highly favorable trade-off between clarity and efficiency.

The method results in Table~\ref{tab:exp-methods} demonstrate that simple prompt-based interventions can fundamentally alter a model's underlying clarification policy, actively shifting the trade-off between conversational efficiency and accurate intent resolution. For both Qwen 3.5 Flash and Llama 4 Maverick, the explanation method results in higher whole-dialogue rewards, lower normalized regret, and higher success rate than the baseline, whereas one-shot prompting primarily serves to minimize average turns, interaction costs, and penalties. These results suggest that prompting methods can shift the policy trade-off: some make models more efficient, while others improve intent resolution.

\footnotetext{\label{thinking} Reasoning mode is disabled for the model.}

\begin{table*}
\centering
\resizebox{\textwidth}{!}{
\begin{tabular}{ll|rrrr|rrr}
\hline
\textbf{Model} & \textbf{Persona} & \textbf{Reward} $\uparrow$ & \textbf{Regret} $\downarrow$ & \textbf{N.Regret} $\downarrow$ & \textbf{Success} $\uparrow$ & \textbf{Avg. Turn} $\downarrow$ & \textbf{Avg. Cost} $\downarrow$ & \textbf{Avg. Penalty} $\downarrow$\\
\hline
Qwen 3.5 Flash & Cooperative & \textbf{0.5655} & \textbf{0.4301} & \textbf{0.7456} & \textbf{0.7110} & \textbf{1.0663} & \textbf{0.1136} & \textbf{0.0319} \\
Qwen 3.5 Flash & Vague & 0.4334 & 0.5569 & 0.9573 & 0.6714 & 1.9846 & 0.1876 & 0.0504 \\
Qwen 3.5 Flash & Contradictory & 0.4004 & 0.5883 & 0.9994 & 0.6711 & 2.3275 & 0.2038 & 0.0669 \\
\hline
Llama 4 Maverick & Cooperative & \textbf{0.5649} & \textbf{0.4307} & \textbf{0.7449} & \textbf{0.7746} & \textbf{1.6321} & \textbf{0.1385} & 0.0712 \\
Llama 4 Maverick & Vague & 0.5292 & 0.4611 & 0.7925 & 0.7739 & 2.0263 & 0.1749 & \textbf{0.0697} \\
Llama 4 Maverick & Contradictory & 0.4897 & 0.4989 & 0.8477 & 0.7685 & 2.4681 & 0.1968 & 0.0819 \\
\hline
\end{tabular}
}
\caption{Performance across different users' personas. The best results are in \textbf{bold}.}
\label{tab:exp-persona}
\end{table*}

\begin{table*}
\centering
\small
\setlength{\tabcolsep}{5pt}
\begin{tabular}{lrrrr}
\toprule
\textbf{Model} 
& \textbf{Reward} $\uparrow$ 
& \textbf{N.Regret} $\downarrow$ 
& \textbf{Avg. Turn} $\downarrow$ 
& \textbf{Avg. Cost} $\downarrow$ \\
\midrule
Grok 4.1 Fast & $-$0.0647 & 2.0064 & 2.8683 & 0.5823 \\
Qwen 3.5 Flash & 0.4283 & 1.0738 & \textbf{2.0782} & \textbf{0.2787} \\
Llama 4 Maverick & \textbf{0.4699} & \textbf{0.9932} & 2.3867 & 0.3094 \\
\bottomrule
\end{tabular}
\caption{Product-search interaction results on PSCon-derived CIGs. The best results are in \textbf{bold}.}
\label{tab:pscon_results}
\end{table*}

\subsection{Dataset-level Reward vs. Success Rate}

Success rate, i.e., accuracy, measures whether the model eventually selects the correct intent, but it does not capture how the model gets there. A clarification policy should also be judged by whether it asks useful questions, avoids unnecessary turns, and stops at the right time.

Table~\ref{tab:exp-models} shows that success rate and reward can lead to different conclusions. Llama 4 Scout and K2Think V2 achieve non-trivial success rates but receive much lower rewards due to high costs or penalties. Conversely, GPT 4.1 and DeepSeek V4 Flash achieve strong rewards with low cost and penalty, demonstrating better policy trade-offs.

Table~\ref{tab:exp-methods} shows a similar pattern for prompting methods. One-shot prompting can reduce turns and cost, sometimes at the expense of success, while the explanation method gives the best overall balance for both Qwen 3.5 Flash and Llama 4 Maverick.

Overall, accuracy remains important, but it is not sufficient for evaluating clarification. Dataset-level reward and normalized regret reveal whether a model resolves intent efficiently and with well-grounded interaction.

\subsection{Persona Robustness}

We experiment by evaluating the same setting under three personas: \textit{cooperative}, \textit{vague}, and \textit{contradictory}. These personas vary in how directly and reliably the user responds to questions, as detailed in Appendix~\ref{app:persona}. Results are shown in Table~\ref{tab:exp-persona}.

The results show that clarification quality depends strongly on user behavior. Both models perform best with cooperative users and degrade as responses become vague or contradictory. Qwen 3.5 Flash is concise in the cooperative setting, but its reward drops sharply under harder personas as turns, cost, and penalty increase. This suggests that the model is sensitive to imperfect user replies and spends more time in interaction without recovering sufficient intent certainty. Llama 4 Maverick is more stable with decreasing rewards, but the success rate remains nearly unchanged. This indicates a different trade-off: less concise under cooperative interaction, but more robust when user responses are incomplete or noisy.

These results show that clarification quality depends not only on the model, but also on user behavior. RegretBench captures this by measuring the additional cost and penalty required to maintain intent resolution in less cooperative interactions.

\subsection{Product Recommendation Interaction}

We evaluate RegretBench in a practical product-search setting using the PSCon subset~\citep{zou2025pscon}. Here, an initial user request may match multiple products, so the assistant must clarify preferences such as product type, features, budget, or intended use before making a recommendation. Results are shown in Table~\ref{tab:pscon_results}.

In product search, clarification is part of the decision process: the assistant must identify which product the user actually wants, not simply produce a plausible recommendation. Llama 4 Maverick achieves the best reward and lowest normalized regret, indicating the strongest balance between identifying the intended product and controlling interaction cost.

Qwen 3.5 Flash uses fewer turns and lower cost, but its higher regret shows that brevity alone is not enough. Grok 4.1 Fast performs worst, with negative reward, high regret, and the highest cost, suggesting that its additional interaction does not translate into useful intent resolution.


This experiment shows that RegretBench applies beyond QA-style ambiguity: in product search, good clarification is part of making the right recommendation, not merely a conversational add-on.

\section{Conclusion and Future Work}

We presented RegretBench, a benchmark for evaluating clarification as sequential decision-making under hidden user intent. RegretBench scores the whole dialogue: whether the assistant identifies the intended interpretation, how much interaction it requires, and whether its clarification questions are grounded in useful ambiguity variables.

Our experiments show that final accuracy is necessary but insufficient. Models with similar success rates can differ substantially in reward due to unnecessary turns, unsupported questions, or poor stopping behavior. Persona and product-search results further show that robust clarification requires handling noisy user replies and practical preference-elicitation settings.

In the future, we plan to extend RegretBench to more scenarios, validate simulations with human users, improve calibration of the parser and simulator, and use the reward signal for policy optimization or reinforcement learning. More broadly, RegretBench encourages evaluating conversational assistants as decision-making systems whose value depends on both success and interaction efficiency.
\section*{Limitations}

RegretBench depends on the quality of its semantic clarification interaction graphs. Although we use precision-oriented filtering, some instances may still contain weak latent variables, redundant facets, or ambiguity dimensions that do not perfectly isolate intent. This filter also favors quality over coverage and may exclude cases of subtle ambiguity.

The benchmark also relies on semantic parsing, simulation, and verbalization. These components enable scalable free-form evaluation, but parser or simulator errors can affect scores. A useful clarification may occasionally be marked unsupported, and simulated personas cannot fully capture real human behavior.

Finally, regret is benchmark-relative rather than an oracle measure of all possible dialogue. The reference planner uses the same semantic state space, observation model, costs, and horizon as the runtime, so results should be interpreted within that abstraction. Different applications may also require different reward trade-offs, especially in high-stakes settings where extra confirmation may be preferable to minimal interaction.

\section*{Ethics and Broader Impact}

\paragraph{Data Collection and Licenses.}
We comply with the terms of all source data licenses and retain the required copyright, license, and attribution notices. AmbigDocs~\cite{ambigdocs} and CondAmbigQA-2K~\cite{li-etal-2025-condambigqa} are released under Apache-2.0 license, and PSCon~\cite{zou2025pscon} under CC-BY-4.0 license. Since RegretBench is derived from these sources, we release our dataset under the CC-BY-4.0 license. 

\paragraph{Security Implications.}
RegretBench aims to improve conversational assistants by testing whether they ask useful clarification questions before answering ambiguous requests. This can reduce misunderstanding and improve reliability. However, strong performance on RegretBench should not be treated as a full safety guarantee, especially in high-stakes domains such as medicine, law, or finance. Moreover, we manually and automatically inspected retained instances for sensitive personal information and offensive content. Since we do not collect new user data, we retain only task-relevant public entity names and remove examples containing private personal details or content not needed for ambiguity evaluation.

\paragraph{Responsible Use.}
RegretBench is an evaluation tool, not a complete measure of model quality. Its scores reflect benchmark-specific rewards, costs, and penalties. Results should be interpreted together with human evaluation, error analysis, and domain-specific testing.

\paragraph{Bias and Fairness.}
Systems may affect users differently depending on their background, writing style, or knowledge. RegretBench uses free-form dialogue and simulated personas, but it may still inherit biases from source datasets or simulators. 

\paragraph{Transparency and Reproducibility.}
We support open and reproducible research by releasing our benchmark under open-source licenses\footnote{\url{https://github.com/ngocminhta/RegretBench}}.

\bibliography{ref}
\bibliographystyle{acl_natbib}

\appendix
\clearpage
\section*{Appendix}
\appendix

\section{Dataset Filtering and Quality Control}
\label{app:dataset_filtering}

RegretBench evaluates whether an assistant can recognize ambiguity, ask a useful clarification question, and answer once the intended referent is identified. This requires stricter filtering than simply keeping records with multiple documents or answers. At test time, the assistant sees only the user question and dialogue history; it does not see entity annotations, document titles, snippets, candidate answers, or hidden intent labels. Therefore, the ambiguity must be recognizable from the user-facing question itself.

We apply a precision-oriented filter before converting source records into CIGs. A retained example should describe a realistic self-clarification scenario: given only the user question, a competent assistant would have a reasonable basis for asking which entity, version, jurisdiction, year, product, or interpretation the user means. Source documents are used only after this question-level check, to verify that the candidate referents are genuine and that their answers are supported. They are not used to justify keeping examples whose ambiguity is invisible from the question alone.

\subsection{Filtering Principles}

We keep records that satisfy four main conditions. First, the ambiguity should be salient from the question text. This often occurs when the ambiguous expression belongs to a class with recurring names, such as films, books, songs, laws, elections, political parties, forts, temples, airports, landmarks, or places. In such cases, clarification by year, location, jurisdiction, medium, or entity type is natural.

Second, the ambiguous expression should not have an overwhelmingly dominant default referent. Some names are technically ambiguous but strongly associated with one meaning. In those cases, asking for clarification would be unnatural, and the example mostly tests whether a model avoids a reasonable default rather than whether it recognizes genuine ambiguity.

Third, the candidate referents must be clean alternatives for the same surface form. They should not be misspellings, redirects, fragments, loose near-matches, or unrelated entities. The alternatives should be distinguishable with concise labels, such as country, city, year, domain, role, occupation, media type, or product attribute.

Fourth, the ambiguity should matter for the answer. After the user selects a referent, the final answer should be directly supported by the corresponding source document and should differ meaningfully across alternatives. If clarification does not change the answer, the record is not useful for RegretBench.

\subsection{Common Drop Cases}

We drop records that fail the question-level ambiguity test or have poor candidate structure. Common examples include generic biography questions about obscure people, such as asking for an unknown person's profession, spouse, or office date, when the name itself has only a single, reasonably salient referent. These examples mainly test obscure entity knowledge rather than clarification.

We also drop many small-place census or administrative lookups, especially questions about the population or location of villages, townships, gminas, census-designated places, or local administrative units. Such records are kept only when the place name has clean, independently plausible alternatives with clear labels; otherwise, the ambiguity is usually visible only from the hidden documents.

Another major category is hidden-document ambiguity. Here, the documents reveal multiple possible entities, but the user's question itself provides no clear indication that clarification is needed. These records are excluded because the evaluated assistant would not see the documents at test time.

Finally, we remove records with unnatural wording, bad candidate sets, unsupported or messy answers, too many hard-to-distinguish alternatives, or questions that already specify the intended referent. For borderline cases, we prefer dropping the record. The filter prioritizes quality and realism over coverage.

\subsection{Examples}

Table~\ref{tab:filtering_examples} gives representative examples of the filtering decisions. The filter is not simply based on fame or popularity: niche facts can be retained when the ambiguity is natural, the alternatives are cleanly separable, and a short clarification could resolve them. Conversely, a technically ambiguous question is dropped if the ambiguity is only visible from hidden annotations, metadata, or source-specific evidence rather than the user-facing request.

\begin{table*}
\centering
\small
\begin{tabular}{p{0.30\linewidth}p{0.08\linewidth}p{0.54\linewidth}}
\toprule
\textbf{Question} & \textbf{Decision} & \textbf{Rationale} \\
\midrule
What was the profession of Edward Andrews? & DROP 
& This is a generic biography question about a personal name. Unless the candidate referents are independently salient, the task mainly tests obscure-person recognition rather than clarification behavior. \\

What was Joseph Fox's profession? & DROP 
& The ambiguity is not naturally cued by the question. A normal assistant would not have a strong reason to ask for clarification from the question alone. \\

What is the population of Rejowiec? & DROP 
& This is a small-place census or administrative lookup. The ambiguity is likely visible only through hidden documents rather than the user-facing question. \\

What is Neymar known for? & DROP 
& One referent is overwhelmingly dominant. Asking for clarification would be unnatural unless the user provided another cue. \\

What did Bernoulli discover? & KEEP 
& The family name is a natural ambiguity cue when candidate documents correspond to multiple notable Bernoulli scientists. Clarifying which Bernoulli the user means is reasonable. \\

Where is Fort Amsterdam located? & KEEP 
& Fort and landmark names often have multiple instances. The alternatives can be cleanly distinguished by country or location, and the final answer depends on the selected referent. \\

When was the United Liberal Party formed? & KEEP 
& Political parties with the same name may exist in different countries or historical contexts. Clarification by country or jurisdiction is natural. \\

Who are the main actors in the film Red Dust? & KEEP 
& Multiple films can share the same title. Asking which version or year the user means is a useful clarification. \\

What does the Access to Justice Act deal with? & KEEP 
& Laws with similar names can exist across jurisdictions or versions. Clarifying the jurisdiction or act version is natural and affects the answer. \\
\bottomrule
\end{tabular}
\caption{Representative filtering examples. Kept records are those in which the ambiguity is evident from the user question and can be resolved with a concise clarification question.}
\label{tab:filtering_examples}
\end{table*}




\section{Supporting Metrics and Diagnostics}
\label{app:supporting_metrics}

\paragraph{Average turns and average total cost.}
Average turns measures the mean number of clarification turns before the final answer or terminal commitment. Average total cost reports the mean interaction cost under the active cost profile. In the current token-based setting, this primarily reflects assistant-side token usage. These metrics measure efficiency. They help distinguish models that resolve ambiguity concisely from those that rely on long or repetitive interaction.

\paragraph{Average total penalty.}
Average total penalty reports the mean penalty accumulated from unsupported, noisy, redundant, or incomplete interactions. It captures interaction quality beyond correctness. A high penalty suggests the model's clarification strategy is poorly grounded, brittle under user variability, or unable to stop cleanly.

\paragraph{Unsupported clarification rate.}
The unsupported clarification rate measures how often the assistant asks a question that cannot be mapped to a supported semantic clarification action. This is an important diagnostic for free-form evaluation. A question may sound natural, but fail to target a latent variable distinguishing candidate intents. A high rate usually indicates vague, answer-seeking, multi-facet, or ungrounded clarification behavior.

\paragraph{Forced terminal answer rate.}
The forced terminal answer rate measures how often the benchmark must force a final commitment because the model reaches the maximum horizon without producing an answer. This metric diagnoses failure to stop. A high value suggests that the model may continue clarifying after sufficient information has been obtained, or may struggle to transition from asking questions to committing to an intent.

\section{Clarification Methods}
\label{app:clarification_methods}

We evaluate two methods for improving clarification behavior. Both keep the model, runtime, simulator, reward, and configuration unchanged, differing only in extra prompt instructions. The goal is to test whether simple guidance improves sequential clarification without training or external tools.

\paragraph{One-shot.}
It adds a single best-practice clarification dialogue to the model prompt. The example demonstrates the desired interaction pattern: ask one concrete disambiguating question, wait for the user's reply, then answer once the missing variable has been resolved. The prompt also explicitly discourages common failure modes, including repeated clarification, asking for the final answer directly, and listing all possible meanings before asking a follow-up question.

A simplified example is shown below:

\begin{quote}
\textbf{User:} ``Where is the City-County Building?'' \\
\textbf{Assistant:} ``Which city or location do you mean?'' \\
\textbf{User:} ``Salt Lake City.'' \\
\textbf{Assistant:} ``The City-County Building in Salt Lake City is the Salt Lake City and County Building.''
\end{quote}

\paragraph{Explanation.}
This method adds concise decision guidance describing when the assistant should ask a clarification question and when it should answer directly. The guidance also includes short examples illustrating both good and bad clarification behavior. 

\begin{quote}
\textbf{Example 1 -- Unresolved ambiguity:}
User asks: "Where is the City-County Building?"
The assistant should ask one concrete missing facet, not answer all possibilities.
Return: \{"type": "ASK", "question\_text": "Which city or location do you mean?"\}

\textbf{Example 2 -- Resolved ambiguity:} Conversation says the user means Salt Lake City.
The assistant should answer now instead of confirming Salt Lake City again.
Return: \{"type": "ANSWER", "final\_response": "The City-County Building in
Salt Lake City is the Salt Lake City and County Building."\}

\textbf{Example 3 -- Bad clarification to avoid:}
Do not ask "Which person do you mean?" if a more concrete facet, such as
occupation, domain, era, city, or work, is available.
\end{quote}

\section{Personas}
\label{app:persona}

RegretBench uses persona-conditioned user simulation to evaluate whether clarification policies are robust to different user behaviors. A persona specifies how the benchmark-side user simulator responds when the assistant asks a clarification question. The same CIG and hidden intent can therefore produce different interaction patterns depending on the persona. 

Personas are used only to control the simulated user. They do not change the candidate intents, latent variables, reward function, reference planner, or evaluation metrics.

This design lets us separate two aspects of performance: whether a model can identify the correct intent under ideal interaction, and whether its clarification policy remains stable when user replies are incomplete, vague, or noisy.

\begin{table}
\centering
\small
\begin{tabular}{p{0.33\linewidth}p{0.56\linewidth}}
\toprule
\textbf{Persona} & \textbf{Behavior} \\
\midrule
Cooperative 
& The user tends to answer valid clarification questions directly and provides the missing information when it is available. This persona measures performance under relatively clean interaction. \\

Vague 
& The user may provide partial, underspecified, or hesitant replies. This persona tests whether the assistant can recover from incomplete clarification signals without over-asking. \\

Contradictory 
& The user may occasionally provide conflicting or noisy evidence. This persona tests whether the assistant can remain robust when user replies are less reliable. \\
\bottomrule
\end{tabular}
\caption{User personas used by the benchmark-side simulator. Personas affect the behavior of simulated user replies, not the underlying hidden intent or CIG structure.}
\label{tab:personas}
\end{table}

Each persona, as described in Table~\ref{tab:personas}, is implemented as a structured simulator specification controlling properties such as cooperation, vagueness, contradiction, patience, and verbosity. Cooperation affects how directly the user answers a valid clarification question. Vagueness controls how often the user omits details or remains underspecified. Contradiction controls the amount of noisy or conflicting evidence. Patience affects tolerance for repeated or inefficient clarification, and verbosity controls the surface form of user replies.

\end{document}